%% file: main.tex
\documentclass[11pt,logo]{techreport}

\usepackage{mathtools}
\usepackage{multirow}
\usepackage{subcaption}
\usepackage[capitalize,noabbrev]{cleveref}

\title{Intern-Med: Step-Level Credit Assignment for Medical Reasoning}
\shorttitle{Intern-Med Technical Report}

\author[1]{Shuyun Yang\textsuperscript{*}}
\author[1]{Rongzhao Zhang\textsuperscript{*}}
\author[1,2]{Tao Yang\textsuperscript{*}}
\author[1]{Yiran Wang}
\author[1,3]{Xiangyu Li}
\author[1]{Lantian Zhang}
\author[1]{Lu Lu}
\author[1]{Yun Zhong}
\author[1]{Jinru Ding}
\author[1]{Mouxiao Bian}
\author[1]{Yi Fu}
\author[1]{Zhao Jin}
\author[1]{Ming Hu}
\author[1]{Yirong Chen}
\author[1]{Jie Xu\textsuperscript{$\dagger$}}

\affil[1]{Shanghai AI Laboratory}
\affil[2]{Shanghai Jiao Tong University}
\affil[3]{Tianjin University}
\affil[*]{\normalsize Equal contribution}
\affil[$\dagger$]{\normalsize Corresponding author: \texttt{xujie@pjlab.org.cn}}

\reportdate{}
\reportstatus{Public draft}
\correspondingauthor{}

\reportlogo{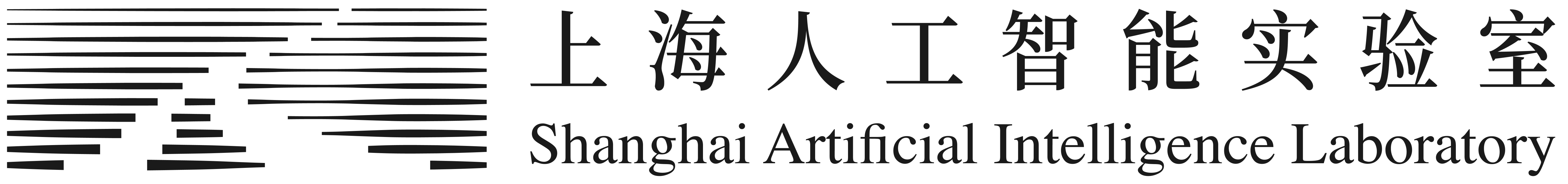}
\reportlogowidth{140pt}

\reportabstract{%
Clinical reasoning is inherently structured: physicians gather evidence, form hypotheses, compare alternatives, and commit to decisions under uncertainty. Yet most approaches to medical specialization of large language models treat reasoning as flat text, supervising and optimizing it only at the trajectory level. We present Intern-Med, a medicine-specialized language model initialized from Intern-S2-Preview and developed through a progressive pipeline that brings structure to both supervision and optimization. Rather than modifying the underlying architecture, we construct knowledge-complete, step-delimited training data at scale, and optimize medical reasoning with MedState-GPO, a reinforcement learning algorithm that tracks the reasoning state of sampled responses against prompt-specific clinical checkpoints and assigns step-level credit according to each step's contribution to reasoning progress. On MedBench, a comprehensive benchmark for medical capability, Intern-Med attains a score of 62.4, a 10.6\% relative improvement over its base model (56.4); as of November 2026, this score would place fourth on the official MedBench open-source leaderboard for large language models. On six additional medical benchmarks, it reaches an average accuracy of 84.1\%, comparable to or exceeding the external models across the board. On general-purpose and math benchmarks, Intern-Med remains comparable to its base model, with a remaining gap to external models on MATH. These results indicate that explicitly structured supervision and optimization can deliver strong medical specialization without compromising general competence. This technical report details the full development process of Intern-Med. By sharing our practical insights, we aim to establish a data-centric and structure-aware development methodology that advances healthcare AI.
}

\begin{document}

\maketitle

\input{sections/01-introduction}
\input{sections/02-data-engineering}

\input{sections/03-cold-start}
\input{sections/04-post-training}

\input{sections/05-evaluation}

\input{sections/06-model-card}

\bibliographystyle{plainnat}
\bibliography{refs}

\end{document}

%% file: sections/01-introduction.tex
\section{Introduction}
\label{sec:introduction}

Large language models (LLMs) have demonstrated strong general capabilities in language understanding, knowledge-intensive reasoning, and instruction following, providing a flexible foundation for adaptation to specialized domains~\cite{deepseekr1}. In medicine, however, directly applying general-purpose models remains challenging. Medical tasks require not only broad factual knowledge, but also precise terminology, structured clinical reasoning, and reliable use of domain-specific evidence. These requirements are difficult to satisfy through limited instruction tuning alone, particularly when medical training data are heterogeneous in format, uneven in quality, and frequently dependent on implicit or missing contextual knowledge.
\begin{figure}[h]
    \centering
    \includegraphics[width=1.0\linewidth]{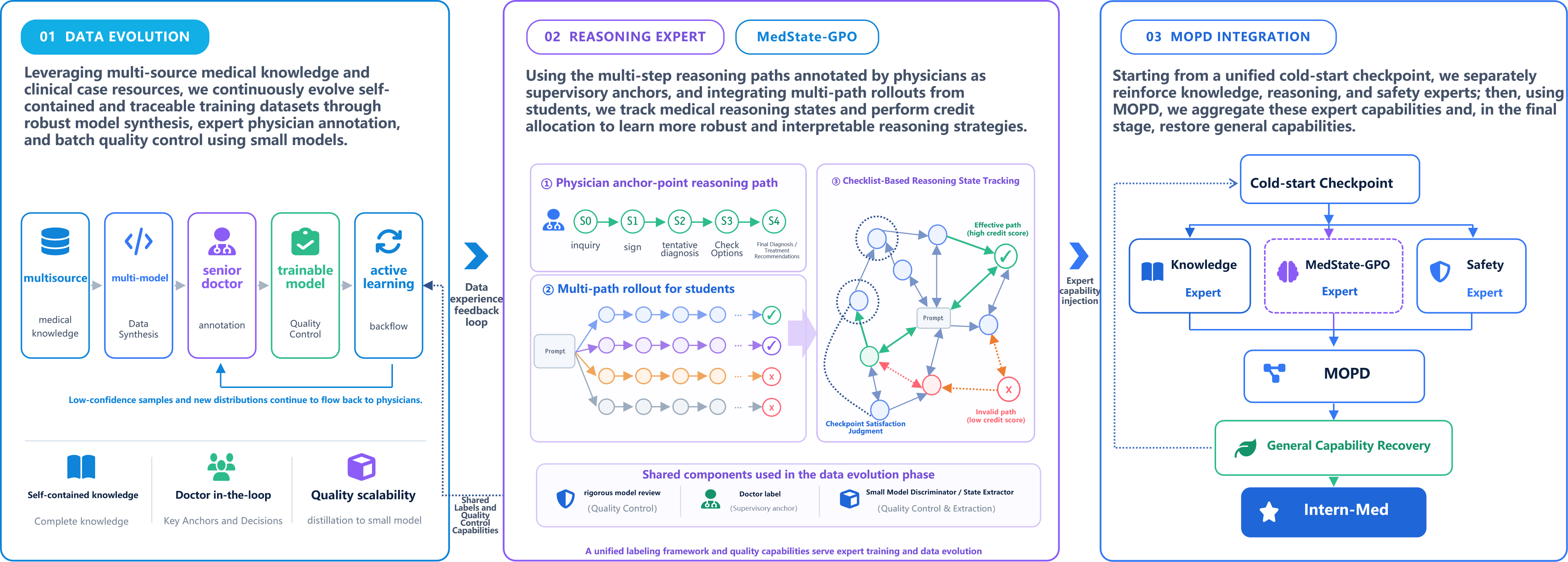}
    \caption{Overview of the Intern-Med development pipeline. The framework consists of three key stages: (1) a data pipeline that curates and processes large-scale medical corpora for training; (2) multi-step reasoning reinforcement learning, which enhances the model's clinical reasoning capabilities through iterative reward-driven optimization; and (3) the MOPD (Multi-teacher On-Policy Distillation) process, which transfers knowledge from multiple expert teachers into a unified student model. Together, these stages form an end-to-end workflow for building Intern-Med.}
    \label{fig:overview}
\end{figure}
Beyond data quality lies a more fundamental gap between how clinicians reason and how models are trained. Clinical reasoning is inherently structured: physicians gather evidence, generate and rank hypotheses, eliminate alternatives, and commit to decisions under uncertainty~\cite{amie}. Mainstream training pipelines, by contrast, treat reasoning as flat text. Supervised fine-tuning imitates final answers or unsegmented rationales, and outcome-based reinforcement learning~\cite{deepseekmath,deepseekr1} assigns identical credit to every step of a trajectory, whether it is a sound evidence-extraction step or an erroneous differential judgment. We argue that medicine is therefore not merely an application domain for large language models, but a natural testbed for \emph{structured machine reasoning}: when supervision and optimization are organized around the structure of clinical reasoning, medical specialization becomes a controllable engineering process rather than an empirical art.

Recent efforts in medical LLM development have emphasized domain adaptation through continued pre-training, supervised fine-tuning, and preference- or reasoning-oriented post-training~\cite{medpalm,huatuogpt,huatuogpt2,baichuanm1,medgemma,lingshu}. While these approaches have improved model performance across medical benchmarks, the effectiveness of domain adaptation depends strongly on the quality and organization of the training data. In practice, medical corpora are collected from diverse sources, including textbooks, guidelines, examination questions, clinical documents, and synthetically generated samples. Differences in terminology, task formulation, annotation granularity, and contextual completeness can introduce substantial noise into training. Moreover, many samples assume external background knowledge or omit intermediate information required to derive the target answer, making the supervision itself incomplete. Under such conditions, simply increasing the amount of domain data does not necessarily translate into more effective knowledge acquisition.

Intern-Med is developed from a different perspective: we treat medical specialization primarily as a supervision construction problem, rather than an architectural one. Starting from Intern-S2-Preview, we retain the original model architecture and build a progressive mid-training and post-training pipeline around a unified medical data framework. A central objective of this framework is to normalize heterogeneous medical resources into consistent training formats and apply a unified cleaning process across different stages of model development. This allows data from different sources and task types to be incorporated under a common quality standard, while preserving their domain-specific information.

More importantly, we explicitly consider knowledge completeness at the data construction stage. During synthetic data generation, we aim to incorporate the domain knowledge necessary for solving each task directly into the corresponding training instance, so that the supervision is as self-contained as possible. Rather than relying on the model to recover missing premises from its pretrained knowledge, each sample is constructed to provide sufficient context for understanding the problem, performing the required reasoning, and producing the expected response. This principle is applied across data synthesis, normalization, filtering, and subsequent training, with the goal of improving both the density of useful supervision and the traceability of knowledge introduced during domain adaptation.

Intern-Med operationalizes this view of structured reasoning at three levels. At the supervision level, heterogeneous medical resources are normalized into consistent, knowledge-complete training formats, and multi-step reasoning chains are synthesized with explicitly delimited steps, so that the reasoning process itself becomes inspectable and segmentable (Section~\ref{sec:data}). At the optimization level---the core algorithmic contribution of this report---we introduce MedState-GPO, a reinforcement learning algorithm that tracks the reasoning state of sampled responses against prompt-specific clinical checkpoints and assigns step-level credit according to each step's contribution to reasoning progress, without requiring a critic or a process reward model (Section~\ref{sec:post-training:medstate}). At the system level, capability-specific specialist branches are trained from a shared cold-start checkpoint and consolidated into a single model through Multi-teacher On-Policy Distillation (MOPD), so that the contribution of each capability can be attributed, combined, and audited independently (Section~\ref{sec:post-training:mopd}).

Intern-Med represents an initial exploration of this direction. We develop the model through progressive mid-training and post-training, and evaluate it on MedBench~\cite{medbench} and a suite of public benchmarks. Beyond reporting benchmark performance, this report documents the development process itself, including the construction and normalization of medical data, synthetic data generation, unified quality control, and the organization of different training stages. We view structured reasoning not as a solved problem, but as a promising direction for medical AI: our results suggest that knowledge-complete supervision, combined with optimization that respects the structure of reasoning, offers a practical path toward medical specialization without compromising general competence. We hope this exploration provides a useful reference for the community, and we look forward to further investigation into structure-aware training for high-stakes domains.

%% file: sections/02-data-engineering.tex
\section{Data Engineering}
\label{sec:data}

Data construction is the foundation of Intern-Med. We build the training corpus in three steps: collecting and cleaning data from public and private sources, measuring the quality of every sample by its actual effect on downstream medical performance, and synthesizing structured multi-step reasoning chains as the basis for cold-start training (Section~\ref{sec:data:cot}).

\subsection{Data Curation}
\label{sec:data:sources}

The corpus originates from publicly released datasets and in-house private medical data, with general-domain data sampled from public datasets to maintain general competence alongside medical specialization. All data are converted into a unified conversational format and cleaned by rule-based procedures only, without any model inference: format and encoding validation, removal of empty, truncated, or garbled samples, language identification, repetition filtering, exact and near-duplicate elimination over normalized text, and benchmark decontamination against MedBench, the test splits of public benchmarks, and our private held-out test set.

Heuristic quality scores do not necessarily reflect whether a sample improves a medical model, so we assess quality by measured downstream utility. The cleaned corpus is partitioned into candidate subsets by source and task type, and each subset is evaluated through controlled proxy-training experiments: a small model is fine-tuned with and without the subset under an identical configuration, and the subset is labeled high- or low-quality according to the resulting change on a private held-out test set that has never been exposed to training or public release. These experimentally grounded labels supervise a quality classifier that scores the full corpus. Samples the classifier judges with high confidence are retained or discarded directly, while uncertain samples are reviewed by senior physicians; their labels are fed back to retrain the classifier, and this physician-in-the-loop iteration continues so that expert effort concentrates on the samples the model finds hardest to judge.

The data retained after the final iteration constitute the training corpus, in which medical data are mixed with general and mathematical data. The medical portion spans six task categories of increasing complexity—selection, extraction, classification and verification, normalization, reasoning-based question answering and decision making, and language generation with medical-insurance scenarios—and six knowledge domains, with basic medicine and the clinical specialties forming the backbone, complemented by public health, stomatology, evidence-based medicine, and clinical informatics. Retained general-domain data are replayed in later training stages to preserve general capabilities.

\subsection{Structured Reasoning Chain Synthesis}
\label{sec:data:cot}

The final component of the data pipeline supplies the corpus for cold-start training (Section~\ref{sec:cold-start}). Cold-start training performs the initial adaptation of the general-purpose base model to the medical domain, and it serves two further purposes that shape the design of the data synthesized here. First, medical reasoning must be shaped into a consistent, mechanically segmentable structure, so that MedState-GPO can later track reasoning states over well-delimited step boundaries and assign credit at the step level during post-training (Section~\ref{sec:post-training:medstate}). Second, because every specialist branch in post-training starts from this same cold-start checkpoint, their token-level distributions remain directly comparable, which allows MOPD to consolidate them into a single model (Section~\ref{sec:post-training:mopd}).

We select prompts that demand multi-step medical reasoning, such as diagnosis, differential comparison, and treatment decisions. A routing model first classifies each prompt into one of eleven reasoning strategies---option selection, clinical diagnosis, differential diagnosis, rule audit, quantitative calculation, quality audit, management planning, prognosis, triage, knowledge explanation, and communication generation---which determines the reasoning pattern to follow. A strong teacher model then generates two candidate chains per prompt under complementary instructions~\cite{star,rft}, one following an evidence-first, minimally sufficient path and the other performing an independent re-examination of the first intuition. Every chain is enclosed in \texttt{<think>} tags and organized into explicitly delimited steps, each performing a single operation such as evidence extraction, hypothesis generation, or differential comparison; following the knowledge-completeness principle (Section~\ref{sec:introduction}), each chain explicitly states the medical knowledge its inferences rely on, and closes with the final answer. Candidates are filtered for format validity and consistency with the reference answer, and the better candidate is selected by multi-sample voting with a selector model.

The verified reasoning chains produced by this pipeline form $\mathcal{D}_{\mathrm{basic\text{-}reasoning}}$, which, together with $\mathcal{D}_{\mathrm{knowledge}}$, constitutes the cold-start corpus $\mathcal{D}_{\mathrm{cold}}$, while medical reasoning prompts with verifiable answers form the reinforcement learning pool $\mathcal{D}_{\mathrm{reason}}$.

%% file: sections/03-cold-start.tex
\section{Cold-Start Training}
\label{sec:cold-start}

Cold-start training~\cite{deepseekr1} transforms the general-purpose initialization into a common medical checkpoint $M_{\mathrm{cold}}$ with broad medical knowledge, standardized medical response behavior, and a fundamental step-wise reasoning pattern. This stage is deliberately positioned before specialist optimization: it provides a shared capability foundation, so that later specialists can be compared and integrated without being confounded by different starting points.

\subsection{Training Data Composition}

The cold-start corpus combines the two strata produced by the data pipeline (Section~\ref{sec:data}):
\begin{equation}
\mathcal{D}_{\mathrm{cold}}
=
\mathcal{D}_{\mathrm{knowledge}}
\cup
\mathcal{D}_{\mathrm{basic\text{-}reasoning}}.
\end{equation}
The knowledge stratum provides broad medical coverage---terminology, mechanisms, diagnostic criteria, and treatment knowledge---while the reasoning stratum contains the synthesized chains that connect evidence, intermediate judgments, and conclusions through an explicit step structure.

Training proceeds in two phases. In the main phase, the model is trained on the full mixture, with sampling controlled across source type, medical specialty, task format, answer length, difficulty, and risk level; high-volume but repetitive task categories are down-sampled, and underrepresented yet clinically relevant domains are up-weighted. In a subsequent annealing phase, the mixture shifts toward the synthesized reasoning chains, consolidating the step-delimited reasoning format in scenarios that require multi-step medical inference. This format is the structural foundation on which MedState-GPO later tracks reasoning states (Section~\ref{sec:post-training:medstate}).

\subsection{Supervised Cold-Start Objective}

Given an input sequence $x$ and a target sequence $y=(y_1,\ldots,y_T)$, the model is optimized with a masked autoregressive objective:
\begin{equation}
\mathcal{L}_{\mathrm{cold}}
=
-
\mathbb{E}_{(x,y)\sim\mathcal{D}_{\mathrm{cold}}}
\left[
\frac{1}{\sum_{t=1}^{T}m_t}
\sum_{t=1}^{T}
m_t
\log
\pi_{\theta}
\left(
y_t \mid x,y_{<t}
\right)
\right],
\end{equation}
where $m_t$ restricts optimization to the response tokens.

Although the objective has the form of standard supervised fine-tuning, its role differs from narrow task adaptation. Because the training data expose the medical context, relevant knowledge, and reasoning premises behind each target response, the model learns to associate medical evidence with domain knowledge, intermediate judgments, and response structures rather than to memorize isolated labels. We preserve variation in response style and task formulation to avoid over-constraining the policy to a single template, while normalizing medical terminology, uncertainty expressions, and safety-critical response patterns wherever inconsistent formatting would otherwise create spurious supervision.

\subsection{Checkpoint Selection and Transition to Post-Training}

Medical and general-domain evaluations are monitored throughout cold-start training. The goal at this stage is not to fully optimize final model behavior, but to obtain a stable medical foundation with sufficient knowledge coverage and a reliable step-wise reasoning pattern; we therefore select the cold-start checkpoint by jointly considering medical capability, training stability, and general-capability regression.

The selected checkpoint $M_{\mathrm{cold}}$ serves as the common initialization for all specialist branches in post-training (Section~\ref{sec:post-training}). A shared initialization brings two benefits: improvements of individual specialists can be attributed directly to their specialist optimization, and all specialists remain architecturally and tokenization-compatible, which is a prerequisite for token-level MOPD consolidation. Starting from $M_{\mathrm{cold}}$, the knowledge specialist is further optimized on the knowledge stratum, the medical reasoning specialist is trained with MedState-GPO on $\mathcal{D}_{\mathrm{reason}}$, and the safety specialist with preference optimization; the resulting specialists are then consolidated into a single model through MOPD, followed by a general-capability recovery stage.

%% file: sections/04-post-training.tex
\section{Post-Training}
\label{sec:post-training}

Post-training starts from the common cold-start checkpoint $M_{\mathrm{cold}}$ and proceeds in three phases. First, several specialist branches are optimized independently, each targeting a capability that benefits from a dedicated data pool and training objective. Second, the specialists are consolidated into a single student model through Multi-teacher On-Policy Distillation (MOPD). Third, a general-capability recovery stage replays general-domain data to restore any capability regressed during consolidation. Because all specialists share the same initialization, architecture, and tokenizer, their gains can be attributed to specialist optimization, and their token-level distributions can be distilled directly into one model.

\subsection{Specialist Branches}
\label{sec:post-training:experts}

Three specialist branches $T_k$ are trained independently from $M_{\mathrm{cold}}$, each on the data pool that matches its target capability (Section~\ref{sec:data}).

The \emph{knowledge specialist} consolidates broad medical knowledge and standardized response behavior through further supervised optimization on the knowledge stratum $\mathcal{D}_{\mathrm{knowledge}}$. The \emph{medical reasoning specialist} is trained with MedState-GPO, a reinforcement learning algorithm with step-level credit assignment introduced in Section~\ref{sec:post-training:medstate}, on $\mathcal{D}_{\mathrm{reason}}$, a pool of medical reasoning prompts with verifiable answers selected beyond the cold-start corpus. The \emph{safety specialist} is trained with direct preference optimization~\cite{dpo} to refuse unsafe requests and to keep medical responses within appropriate boundaries---declining harmful instructions, acknowledging uncertainty, and avoiding over-commitment in clinical communication.

\subsection{MedState-GPO for Medical Reasoning}
\label{sec:post-training:medstate}

GRPO~\cite{deepseekmath} assigns a single trajectory-level advantage to all tokens of a response, so a sound evidence-extraction step and an erroneous differential judgment in the same response receive identical credit. MedState-GPO obtains step-level credit without a critic or a process reward model~\cite{lightman2023verify,mathshepherd}. Its key idea is to track, for each response, how far the reasoning has progressed along a prompt-specific set of clinically meaningful checkpoints, and to concentrate the outcome-level learning signal on the steps that drive this progression.

\paragraph{Reasoning states.}
For each prompt $x$, we predefine a checklist $\mathcal{C}_x$ of checkpoints that encode the key intermediate conclusions a sound solution is expected to reach, such as essential evidence, necessary differential considerations, and the final decision. Each sampled response is decomposed into steps $(b_1,\dots,b_T)$ according to the step structure established during cold-start training, and a judgment model $\mathcal{J}$ maps each successive step prefix to the subset of checkpoints it has explicitly and correctly established. The reasoning state then evolves monotonically,
\begin{equation}
s_t \;=\; s_{t-1} \;\cup\; \mathcal{J}\big(x,\, b_{\le t}\big), \qquad s_0 = \varnothing,
\end{equation}
from the empty set toward complete coverage of $\mathcal{C}_x$. Monotonicity ensures that confirmed medical facts are not retracted by later text, and clinically decisive details such as laboratory values and dosages are treated with particular care during judgment.

\paragraph{Transition-based credit assignment.}
The progress contributed by step $b_t$ is measured by the number of newly established checkpoints,
\begin{equation}
\Delta_t \;=\; |s_t| - |s_{t-1}| \;\ge\; 0,
\end{equation}
and converted into a step-level weight
\begin{equation}
w_t \;=\; 1 + \alpha \cdot \Delta_t, \qquad \alpha > 0,
\end{equation}
which is shared by all tokens of $b_t$ and normalized within each response so that the mean token weight equals one. A step that advances the reasoning state is therefore up-weighted, while a step that leaves the state unchanged remains neutral. The terminal, verifiable outcome reward $R_i$ of each response in the sampled group $\{y^{(i)}\}_{i=1}^{n}$ is normalized within the group as in GRPO,
\begin{equation}
A^{\mathrm{E}}_i \;=\; \frac{R_i - \operatorname{mean}\big(\{R_j\}_{j=1}^{n}\big)}{\operatorname{std}\big(\{R_j\}_{j=1}^{n}\big)},
\end{equation}
and the token-level advantage combines the two signals,
\begin{equation}
A_{i,t} \;=\; \tilde{w}_{i,t} \cdot A^{\mathrm{E}}_i,
\end{equation}
where $\tilde{w}_{i,t}$ is the normalized weight of the step containing token $t$. The terminal outcome determines the direction and magnitude of the advantage for the response as a whole, while state transitions determine how this advantage is distributed within the trajectory: in successful responses, steps that genuinely acquire new evidence or confirm key judgments receive larger updates, while in failed responses the magnitude of the negative update likewise scales with each step's contribution to progression. The token-level advantages replace the uniform advantage in the clipped surrogate objective~\cite{ppo}. When no sampled response reaches a correct outcome, or states cannot be reliably determined, all step weights equal one and MedState-GPO reduces to GRPO.

\paragraph{Relation to prior work.}
MedState-GPO is related in spirit to graph-based credit assignment~\cite{graphgpo}, which aggregates rollouts into an explicit state-transition graph and credits actions by their reduction of the graph distance to the goal. MedState-GPO instead measures progress directly through checkpoint satisfaction: reasoning states are defined per prompt by clinically meaningful checkpoints rather than by environment observations, and credit follows from state progression rather than from shortest-path distances on a globally merged graph. Behaviors such as raising danger signals or requesting additional information are monitored but not rewarded, because rewarding them unconditionally would encourage clinically unsafe over-alerting.

\subsection{Multi-Teacher On-Policy Distillation}
\label{sec:post-training:mopd}

After specialist training, the specialists $\{T_k\}_{k=1}^{K}$ are consolidated into a unified student $\pi_\theta$, initialized from $M_{\mathrm{cold}}$. Merging the specialist datasets and training with supervised learning would train the student on off-policy text generated by other models; MOPD instead trains the student on its own samples and uses the specialists as token-level teachers~\cite{gkd,minillm}.

Each training prompt $x$ is associated with a teacher index $k(x)$ according to the capability it targets. The student samples a response $y\sim\pi_\theta(\cdot\mid x)$, and the corresponding teacher scores the sampled prefix at every position. The student minimizes the per-token reverse KL divergence to the routed teacher:
\begin{equation}
\mathcal{L}_{\mathrm{MOPD}}
=
\mathbb{E}_{x\sim\mathcal{D}_{\mathrm{MOPD}},\,y\sim\pi_\theta(\cdot\mid x)}
\left[
\frac{1}{|y|}
\sum_{t=1}^{|y|}
\mathrm{KL}\!\left(
\pi_\theta(\cdot\mid x,y_{<t})
\,\big\|\,
\pi_{T_{k(x)}}(\cdot\mid x,y_{<t})
\right)
\right].
\end{equation}
Because the teachers share the tokenizer and architecture of $M_{\mathrm{cold}}$, this dense per-token supervision is well defined without vocabulary alignment. Reverse KL on on-policy samples is mode-seeking and corrects the student on the states it actually visits, which mitigates the compounding errors of off-policy imitation. The prompt mixture $\mathcal{D}_{\mathrm{MOPD}}$ draws from the specialist pools and replays $\mathcal{D}_{\mathrm{general}}$ to limit the regression of general capabilities during consolidation. After consolidation, a lightweight general-capability recovery stage replays general-domain and mathematical data, yielding the final Intern-Med model.

%% file: sections/05-evaluation.tex
\section{Evaluation}
\label{sec:evaluation}

We evaluate Intern-Med on a comprehensive suite of benchmarks, organized around three questions: how it performs on MedBench v5~\cite{medbench}, a comprehensive evaluation of medical capability; how it compares with external models on public medical benchmarks; and whether medical specialization comes at the cost of general or mathematical capability.

\begin{figure}[htp]
    \centering
    \includegraphics[width=0.9\linewidth]{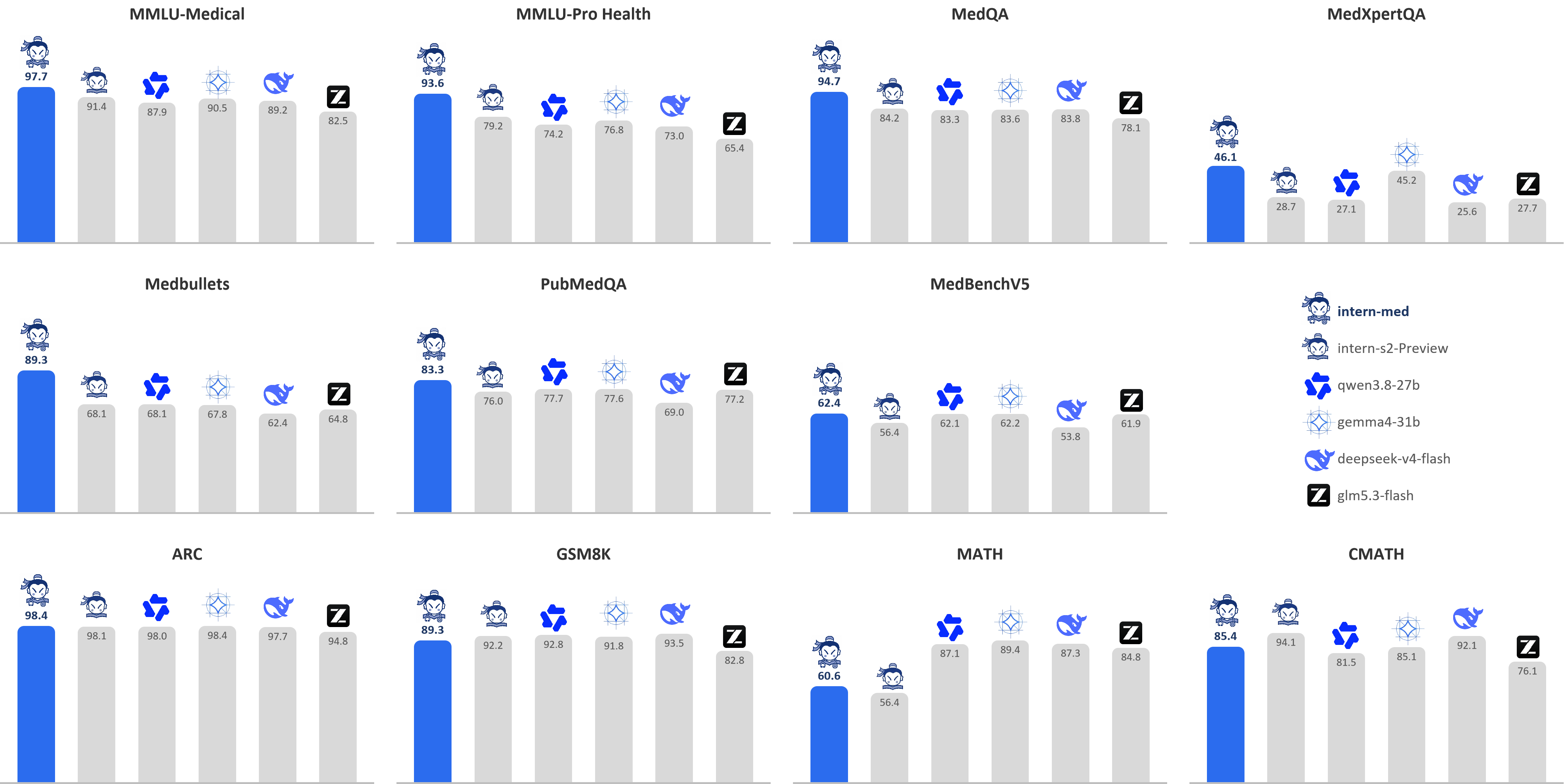}
    \caption{Performance of Intern-Med, its base model Intern-S2-Preview, and external models across medical, general, and math benchmarks.}
    \label{fig:result}
\end{figure}

\subsection{Setup}
\label{sec:evaluation:setup}

Intern-Med is evaluated against its base model Intern-S2-Preview and four external models: two open-weight models of comparable parameter scale, gemma-4-31B and Qwen3.8-27B, and two larger frontier API models, DeepSeek-V4-Flash and GLM-5.3-Flash. Our primary evaluation is MedBench v5, complemented by six public medical benchmarks; general and mathematical capabilities are assessed on ARC, GSM8K, MATH, and CMATH. Accuracy is the reported metric throughout.

\subsection{MedBench v5}
\label{sec:evaluation:medbench}

On MedBench v5, Intern-Med attains a score of 62.4, a 10.6\% relative improvement over its base model (56.4). As of November 2026, this score would place fourth on the official MedBench open-source leaderboard for large language models.

\subsection{Public Medical Benchmarks}
\label{sec:evaluation:medical}

Across the six public medical benchmarks, Intern-Med attains an average accuracy of 84.1\%, an improvement of 12.8 percentage points over the base model Intern-S2-Preview (71.3\%), and exceeds every external model on each benchmark (Figure~\ref{fig:result}). The strongest external model, gemma-4-31B, reaches a medical average of 73.6\%, 10.5 percentage points below Intern-Med. The largest differences appear on Medbullets (89.3\% vs.\ 67.8\%) and MMLU-Pro health~\cite{mmlupro} (93.6\% vs.\ 76.8\%), while the smallest margin is on MedXpertQA~\cite{medxpertqa} (46.1\% vs.\ 45.2\%). Qwen3.8-27B, DeepSeek-V4-Flash, and GLM-5.3-Flash reach medical averages of 69.7\%, 67.2\%, and 66.0\%, respectively, all below Intern-Med.

\subsection{Reasoning-Length-Conditioned Analysis}
\label{sec:evaluation:length}

\begin{figure}[t]
    \centering
    \includegraphics[width=0.8\linewidth]{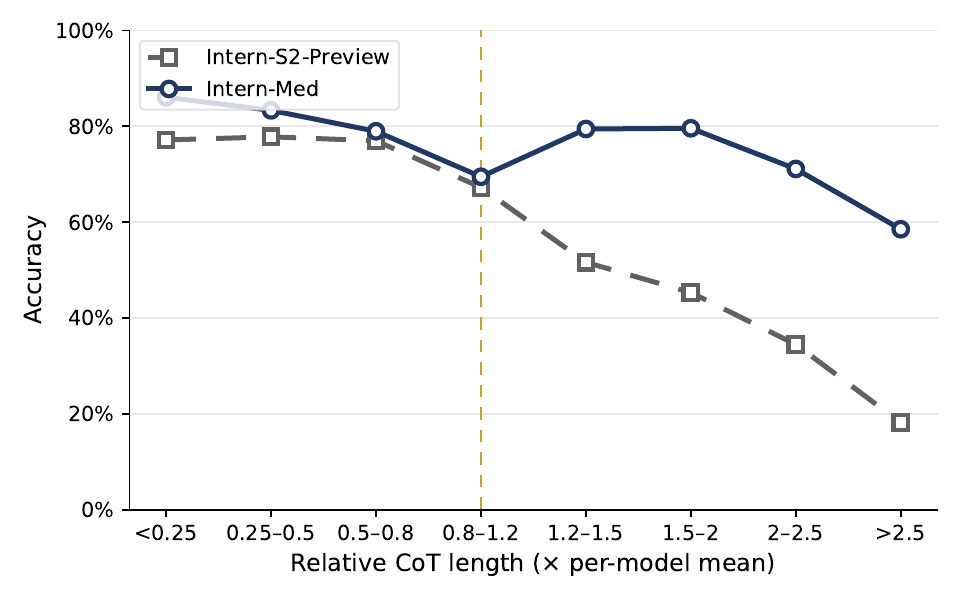}
    \caption{Accuracy conditioned on relative CoT length. Response length is normalized by each model's own mean on the evaluation set (the dashed vertical line marks the mean). Intern-Med maintains accuracy on unusually long reasoning chains, whereas the base model's accuracy degrades monotonically beyond its mean.}
    \label{fig:length}
\end{figure}

To examine how reasoning behavior changes after training, we condition accuracy on the relative length of the generated chain of thought. For each response, we normalize its CoT length by the model's own mean on the evaluation set, group samples into eight relative-length bins, and measure accuracy within each bin.

As shown in Figure~\ref{fig:length}, Intern-Med matches or exceeds its base model in every relative-length bin. The two curves are closest near the per-model mean (relative length $\approx$ 1), where both models sit at about 69\%, while on the short side Intern-Med is already clearly ahead (86\% vs.\ 77\% in the shortest bin). Beyond the mean, the two models diverge monotonically: Intern-Med holds a plateau of 79--80\% for responses between 1.2$\times$ and 2$\times$ its mean length, and still reaches 71\% and 59\% in the 2--2.5$\times$ and $>$2.5$\times$ bins, whereas the base model's accuracy falls from 45\% to 34\% and 18\% over the same range. In other words, when Intern-Med thinks longer than usual---typically on harder problems---its accuracy is largely preserved, while the base model's accuracy collapses as relative reasoning length grows. This pattern is consistent with the design of MedState-GPO: credit is concentrated on steps that advance the reasoning state, so additional reasoning steps are more likely to translate into correct conclusions rather than unproductive drift.

\subsection{Summary}
\label{sec:evaluation:summary}

The results show a clear specialization profile. Intern-Med substantially improves over its base model on MedBench v5 and exceeds all comparison models on every public medical benchmark. This specialization does not compromise broader competence: on the general and math benchmarks, Intern-Med remains comparable to its base model, with 98.4\% on ARC, 89.3\% on GSM8K, 60.6\% on MATH, and 85.4\% on CMATH. A gap to external models remains on MATH, which we leave to future work.

%% file: sections/06-model-card.tex
\section{Model Card}
\label{sec:model-card}

\paragraph{Intended use.}
Intern-Med is a medicine-specialized language model developed for clinical decision support and medical research assistance. It is applicable to tasks such as medical question answering, clinical reasoning over presented cases, medical examination-style problems, and evidence-grounded explanation of medical concepts.

\paragraph{Scope of use.}
Intern-Med is positioned as an assistive tool for healthcare professionals, medical institutions, and researchers. Model outputs must not replace the independent professional judgment of licensed physicians, and should not be used directly for final diagnosis, treatment planning, prescription issuance, or emergency and critical care decisions.

\paragraph{Known limitations and risks.}
\begin{itemize}
\item \textbf{Long-tail and atypical cases.} Due to the coverage of training and evaluation data, the model may be less reliable for rare diseases, atypical presentations, and complex comorbidities.
\item \textbf{Residual hallucination risk.} Although training explicitly targets factual reliability, the model may still produce unsupported statements; critical medical facts should be verified against authoritative sources before use.
\item \textbf{Evaluation coverage.} The reported evaluation is conducted on MedBench and a fixed set of public benchmarks, most of which are in English; performance in other languages, settings, or tasks may differ.
\item \textbf{Knowledge timeliness.} The model's knowledge reflects its training data and may not incorporate the most recent clinical guidelines or regulatory updates.
\item \textbf{Benchmark dynamics.} Reported rankings refer to the MedBench open-source leaderboard as of November 2026 and may change as the leaderboard evolves.
\end{itemize}

\paragraph{Safety.}
A dedicated safety specialist stage is applied during post-training to keep responses within appropriate boundaries: declining harmful instructions, acknowledging uncertainty, and avoiding over-commitment in clinical communication. Behaviors such as raising danger signals are monitored during training. Users should nonetheless treat all outputs as reference information requiring professional review.

\paragraph{Availability.}
The model weights of Intern-Med are publicly available at \url{https://huggingface.co/OpenMedZoo/intern-med}. The training corpus contains a substantial proportion of private medical data and is therefore not publicly released, and training-process details beyond those described in this report are not disclosed.